\documentclass{article}
\usepackage[preprint]{spconf,amsmath,graphicx}
\copyrightnotice{\copyright\ IEEE 2021}
\toappear{To appear in {\it Proc.\ ICASSP2021, June 6-11, 2021, Toronto, Canada}}
\usepackage{amsfonts}
\usepackage{hyperref}
\usepackage{bm}
\usepackage{float}
\usepackage{nicefrac}
\usepackage{support-caption}
\usepackage{subcaption}
\usepackage{multirow}
\usepackage{multicol}
\usepackage{adjustbox}
\usepackage{booktabs}
\usepackage{tablefootnote}
\usepackage{xspace}

\def\X{{\mathbf X}}
\def\x{{\mathbf x}}
\def\H{{\mathbf H}}
\def\h{{\mathbf h}}
\def\W{{\mathbf W}}
\def\a{{\mathbf a}}
\def\M{{\mathbf M}}

\def\espresso{\textsc{Espresso}\xspace}
\def\pychain{\textsc{PyChain}\xspace}

\title{Wake Word Detection with Streaming Transformers}
\name{Yiming Wang$^1$, Hang Lv$^3$, Daniel Povey$^4$, Lei Xie$^3$, Sanjeev Khudanpur$^{1,2}$\thanks{This work was partially supported by an unrestricted gift from \href{https://www.apptek.com/}{Applications Technology (AppTek)}. The authors thank Tongfei Chen and Hainan Xu for valuable comments.}}
\address{
  $^1$ Center for Language and Speech Processing,
  $^2$ Human Language Technology Center of Excellence,\\
  Johns Hopkins University, Baltimore, MD, USA \\
  $^3$ ASLP@NPU, School of Computer Science, Northwestern Polytechnical University, Xi'an, China \\
  $^4$ Xiaomi Corporation, Beijing, China \\
\small{\texttt{\string{yiming.wang,khudanpur\string}@jhu.edu, \string{hanglv,lxie\string}@nwpu-aslp.org, dpovey@gmail.com}}
}
\begin{document}
\ninept
\maketitle
\begin{abstract}
Modern wake word detection systems usually rely on neural networks for acoustic modeling. Transformers has recently shown superior performance over LSTM and convolutional networks in various sequence modeling tasks with their better temporal modeling power. However it is not clear whether this advantage still holds for short-range temporal modeling like wake word detection. Besides, the vanilla Transformer is not directly applicable to the task due to its non-streaming nature and the quadratic time and space complexity. In this paper we explore the performance of several variants of chunk-wise streaming Transformers tailored for wake word detection in a recently proposed LF-MMI system, including looking-ahead to the next chunk, gradient stopping, different positional embedding methods and adding same-layer dependency between chunks. Our experiments on the Mobvoi wake word dataset demonstrate that our proposed Transformer model outperforms the baseline convolution network by 25\% on average in false rejection rate at the same false alarm rate with a comparable model size, while still maintaining linear complexity w.r.t. the sequence length.
\end{abstract}
\section{Introduction}
\label{sec:intro}
\vspace{-1.5mm}
Voice interactions between human and digital assistants integrated in smartphones and home-owned voice
command devices are becoming ubiquitous in our daily lives. This necessitates a built-in wake word
detection system which constantly listens to its environment, expecting a predefined word to
be spotted before turning into a more power consumptive state to understand users' intention (e.g. \cite{wang2019end}). 

Similar to automatic speech recognition (ASR), modern wake word detection systems can be constructed with either HMM/DNN hybrid \cite{panchapagesan2016multi,sun2016compressed,wu2018monophone,wang2020wake}
or pure neural networks \cite{chen2014small,sainath2015convolutional,he2017streaming,shan2018attention,hou2020mining}.
In either of these two categories some types of neural networks are used for acoustic modeling to encode the input features of an audio recording into a high level representation for the decoder to determine whether the wake word
has been detected within some range of frames.

A wake word detection system usually runs on devices, and it needs to be triggered as soon as the wake
word actually appears in a stream of audio. Hence the neural networks are limited to: 1) small memory footprint;
2) small computational cost; and 3) low latency in terms of the number of future frames needed to compute
the score for the current frame. Under these criteria, the family of recurrent neural networks
\cite{hochreiter1997long,cho2014learning} is not suitable because its sequential nature in computation
prevents it from being parallelized across time in the chunk-streaming case even with GPUs. So most of the current systems adopt
convolutional networks. A convolution kernel spans over a small and fixed range of frames, and is
repeatedly applied by sliding across time or frequencies. Although each kernel only captures a local
pattern, the receptive field can be enlarged by stacking several convolution layers together: higher
layers can ``see'' longer range of frames than lower layers, capturing more global patterns.

Recently the self-attention structure, as a building block of the Transformer networks \cite{vaswani2017attention}, receives popularity in both NLP and speech communities for its capability of modeling context dependency for sequence data without recurrent connections \cite{vaswani2017attention,karita2019comparative}. Self-attention replaces recurrence with direct interactions between all the pairs of frames in a layer, making each frame aware of its contexts. The computations are more parallelized, in the sense that the processing of a frame does not depend on the completion of processing other frames in the same layer.  \cite{bai2019time} also explored the self-attention in the keyword search (KWS) task. However, the original self-attention requires the entire input sequence to be available before any frames can be processed, and the computational complexity and memory usage are both $O(T^2)$. Time-restricted self-attention \cite{povey2018time} allows the self-attention to be restricted within a small context window around each
frame with attention masks. But it still does not have a mechanism of saving the current computed 
state for future computations, and thus is not applicable to streaming data. Transformer-XL \cite{dai2019transformer} performs chunk-wise training where the previous chunk is cached as hidden state
for the current chunk to attend to for long-range temporal modeling. So it can be used for streaming tasks. The time and space complexity are both reduced to $O(T)$, and the within-chunk computation across time can be parallelized with GPUs. While there has been recent work \cite{tsunoo2019transformer,tian2020synchronous,zhang2020transformer,lu2020exploring,wu2020streaming} with similar ideas showing that such streaming Transformers achieve competitive performance compared with latency-controlled BiLSTMs \cite{zhang2016highway} or non-streaming Transformers for ASR, it remains unclear how the streaming transformers work for shorter sequence modeling task like wake word detection.

In this paper we investigate various aspects of the streaming Transformers with its application to wake word detection for the recently proposed alignment-free LF-MMI system \cite{wang2020wake}. This paper has the following contributions: 1) we explore how the gradient stopping point during back-propagation affects the performance; 2) we show how different positional embedding methods affect the performance; and 3) we compare the performance of either obtaining the hidden state coming from the current or previous layer. In addition, we propose an efficient way to compute the relative positional embedding in streaming Transformers. To the best of our knowledge, this is the first time that streaming Transformers are applied to this task.

\vspace{-1.5mm}
\section{The Alignment-Free LF-MMI System}
\vspace{-1mm}
We build our system on top of the state-of-the-art system described in \cite{wang2020wake}. We briefly explain that system below to provide some background information. Interested readers can refer to \cite{wang2020wake} for details.

This is a hybrid HMM/DNN system with alignment-free LF-MMI loss \cite{hadian2018end,povey2016purely}, where the positive wake word (denoted as \emph{wake word}) and the negative
non-silence speech (denoted as \emph{freetext}) are modeled with a single left-to-right 4-state HMM respectively,
regardless of how many actual phonemes are there. In addition, a 1-state HMM is dedicated
to model \emph{optional} silence \cite{chen2015pronunciation} (denoted as \emph{SIL}). The motivation behind this
is that we believe the proposed design choice has sufficient modeling power for this task.

In LF-MMI loss, the numerator represents the likelihood of the input feature given the correct output state
sequence, while the denominator represents the likelihood given incorrect state sequences. So the model is trained
to maximize the posterior of the correct sequence among other competing sequences. ``Alignment-free'' here refers to unfolded
FSTs as the numerator graphs are directly derived from the truth labels (``positive'' or ``negative'' in our task). The
denominator graph is specified manually, containing one path corresponding to the positive recordings and two paths
corresponding to the negatives. Since the alignment-free LF-MMI system outperforms the cross-entropy HMM-based and other pure neural systems \cite{wang2020wake}, we base our work in this paper on this specific system.

The work in \cite{wang2020wake} adopts dilated and strided 1D convolutional networks (or
``TDNN''
\cite{peddinti2015time,povey2018semi}) for acoustic modeling, which is straightforward as the
computation of
convolution is both streamable by its nature and highly parallelizable. In the next section, we will
detail our approach to streaming Transformers for modeling the acoustics in our task.

\vspace{-1.5mm}
\section{Streaming Transformers}
\vspace{-1mm}
We recapitulate the computation of a vanilla single-headed Transformer here.\footnote{~A multi-headed extension is straightforward and irrelevant to our discussion here.} Assume the input to a self-attention layer is 
$\X = [\x_1,\ldots, \x_T] \in \mathbb{R}^{d_x \times T}$ where $ \x_j \in \mathbb{R}^{d_x}$.
The tensors of query $\mathbf{Q}$, key $\mathbf{K}$, and value $\mathbf{V}$ are obtained via
\vspace{-1mm}
\begin{equation}
\mathbf{Q}=\W_Q \X, ~~ \mathbf{K}=\W_K \X, ~~ \mathbf{V}=\W_V \X \quad \in \mathbb{R}^{d_h \times T}
\end{equation}
where the weight matrices $\W_Q,\W_K,\W_V \in \mathbb{R}^{d_h \times d_x}$. The output at $i$-th time step is computed as
\vspace{-1mm}
\begin{equation}
\label{eq:compute_val}
\h_i=\mathbf{V} \a_i \in \mathbb{R}^{d_h},\quad \a_i=\mathrm{softmax}\left(\frac{[\mathbf{Q}^\top \mathbf{K}]_i}{\sqrt{d_h}}\right) \in \mathbb{R}^T
\end{equation}
where $[\cdot]_i$ means taking the $i$-th row of a matrix. All the operations mentioned above are 
homogeneous across time, thus can be parallelized on GPUs. Note that here $\mathbf{Q},\mathbf{K},\mathbf{V}$
are computed from the same input $\X$, which represents the entire input sequence. 

\begin{figure}
    \centering
    \subfloat[\centering Dependency on the previous layer of the previous chunk.]{{\includegraphics[width=0.2\textwidth]{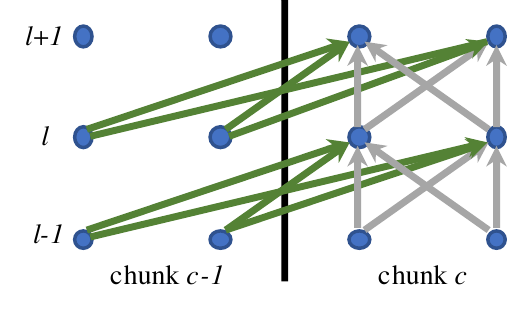} }\label{fig:transformer_dependency_a}}
    \qquad\qquad
    \subfloat[\centering Dependency on the same layer of the previous chunk.]{{\includegraphics[width=0.2\textwidth]{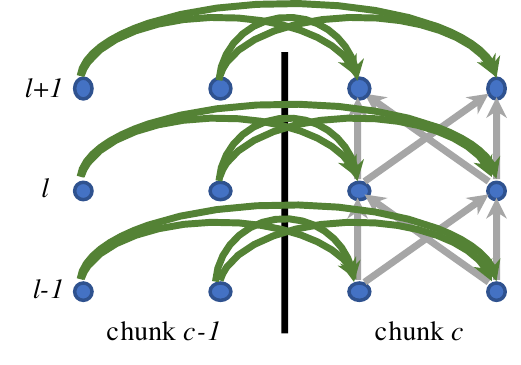} }\label{fig:transformer_dependency_b}}
    \caption{Two different type of nodes dependency when computing self-attention in streaming Transformers. The figures use 3-layer networks with 2 chunks (delimited by the thick vertical line in each sub-figure) of size 2 as examples. The grey arcs illustrate the nodes dependency within the current chunk, while the green arcs show the dependency from the current chunk to the previous one.}
    \label{fig:transformer_dependency}
    \vspace{-4mm}
\end{figure}

Such dependency of each output frame on the entire input sequence makes the model unable to process
streaming data where in each step only a limited number of input frames can be processed. Also,
the self-attention is conducted between every pair of frames within the whole sequence, making the
memory usage and computation cost are both $O(T^2)$. Transformer-XL-like architectures address these concerns by performing a chunk-wise processing of the sequence. The whole input sequence is segmented into several equal-length chunks (except the last one whose length can be shorter). As shown in Fig. \ref{fig:transformer_dependency_a}, the hidden state from the previous chunk are cached to extend keys and values from the current chunk, providing 
extra history to be attended to. In this case, $\tilde{\mathbf{K}}$ (or $\tilde{\mathbf{V}}$) is longer in length than $\mathbf{Q}$ due to the prepended history. To alleviate the gradient explosion/vanishing issue introduced in this kind of recurrent structure, gradient is set to not go beyond the cached state, i.e.,
\vspace{-1mm}
\begin{equation}
\label{eq:state_extended}
\tilde{\mathbf{K}}_c=[\mathrm{SG}(\mathbf{K}_{c-1});\mathbf{K}_c], \quad \tilde{\mathbf{V}}_c=[\mathrm{SG}(\mathbf{V}_{c-1});\mathbf{V}_c]
\end{equation}
where $c$ is the chunk index, $[\cdot;\cdot]$ represents concatenation along the time dimension, and
$\mathrm{SG}(\cdot)$ is the stop gradient operation.\footnote{~For example, this would be \texttt{Tensor.detach()} in PyTorch.} The memory usage and computation cost are both reduced to $O(T)$ given the chunk size is constant.

\vspace{-2mm}
\subsection{Look-ahead to the Future and Gradient Stop in History}
\vspace{-1mm}
\label{sec:look-ahead-and-caching}
Our preliminary experiments show that only having history to the left is not sufficient for a good
performance in our task. So we also allow a chunk to ``look-ahead'' to the next chunk to get future context
when making predictions from the current chunk. For the right context, the gradient in back-propagation does
not just stop at $K_{c+1}$ and $V_{c+1}$, but rather go all the way down to the input within the chunk
$c+1$. On the other hand we can optionally treat the left context (i.e. the history state) the same way as
well. Intuitively, having weights to have more information while being updated should always be beneficial, as long as their gradient flow is constrained within a small range of time steps. This can be achieved by splicing the left chunk together with the current chunk and then only selecting the output of the current chunk for loss evaluation, at the cost of having one more forward computation for each chunk by not caching them. We will show a performance comparison between with and without such state-caching in the experiments.

\vspace{-2mm}
\subsection{Dependency on the Previous Chunk from the Same Layer}
\vspace{-1mm}
\label{sec:same-layer}
Note that when there are multiple stacked self-attention layers, the output of the $c$-th chunk of the $l$-th layer actually depends on the output of the $(c-1)$-th chunk of the $(l-1)$-th layer. So the receptive field of each chunk grows linearly with the number of the self-attention layers, and the current chunk does not have access to previous chunks in the same layer (Fig. \ref{fig:transformer_dependency_a}). This may limit the model’s temporal modeling capability as not all parts of the network within the receptive field are utilized. Hence, we take the output from the previous chunk in the same layer, and prepend it to the key and value. Formally, let 
$\H=[\h_1,\ldots,\h_T] \in \mathbb{R}^{d_h \times T}$ where $\h_i$ is defined in Eq. (\ref{eq:compute_val}). Then Eq. (\ref{eq:state_extended}) becomes:
\vspace{-1mm}
\begin{equation}
\label{eq:state_extended_samelayer}
\tilde{\mathbf{K}}_c^l=[\mathrm{SG}(\H_{c-1}^l);\mathbf{K}_c^l],\quad \tilde{\mathbf{V}}_c^l=[\mathrm{SG}(\H_{c-1}^l);\mathbf{V}_c^l]
\end{equation}
where we use superscript $l$ to emphasize tensors from the same layer. Fig. \ref{fig:transformer_dependency_b} illustrates nodes dependency in such computation flows.

\vspace{-3mm}
\subsection{Positional Embedding}
\vspace{-1mm}
The self-attention in Transformers are invariant to the sequence reordering, i.e., any permutations of the input sequence will yield exactly the same output for each frame. So it is crucial to encode the positions.
The original Transformer \cite{vaswani2017attention} employs deterministic sinusoidal functions to encode absolute positions. In our chunk-streaming setting, we can also enable this by adding an offset to the frame positions within each chunk. However our goal for wake word detection is just to spot the word rather than recognizing the whole utterance, a relative positional encoding may be more appropriate. One type of relative positional embedding, as shown in \cite{shaw2018self}, encodes the relative positions from a query frame to key/value frames in the self-attention, and pairs of frames having the same position difference share the same trainable embedding vector. The embedding vectors $\mathbf{E} \in \mathbb{R}^{d_h\times T}$ are then added to the key (optionally to the value as well) of each self-attention layer. So Eq. (\ref{eq:compute_val}) is modified as:
\vspace{-1mm}
\begin{equation}
\label{eq:compute_val_pos_embed}
\h_i=\left(\mathbf{V}+\mathbf{E}\right) \a_i \in \mathbb{R}^{d_h}, ~~ \a_i=\mathrm{softmax}\left(\frac{[\mathbf{Q}^\top \left(\mathbf{K}+\mathbf{E}\right)]_i}{\sqrt{d_h}}\right) \in \mathbb{R}^T
\end{equation}
As suggested, the relative positional embedding is fed into every self-attention layer and jointly trained with other model parameters.

Different from the case in \cite{shaw2018self} where the query and key (or value) have the same sequence length, there is extra hidden state prepended to the left of the key and the value in the current chunk, making the resulting key and value longer than the query. By leveraging the special structure of the layout of relative positions between the query and key, we design a series of simple but efficient tensor operations to compute self-attentions with positional embedding. Next we show how the dot product between the query $\mathbf{Q}$ and the positional embedding $\mathbf{E}$ for the key $\mathbf{K}$ can be obtained\footnote{~We drop the batch and heads dimensions for clarity. So all tensors become 2D matrices in our description.}. The procedure when adding the embedding to the value $\mathbf{V}$ is similar.

\begin{figure}
  \centering
  \includegraphics[width=0.45\textwidth]{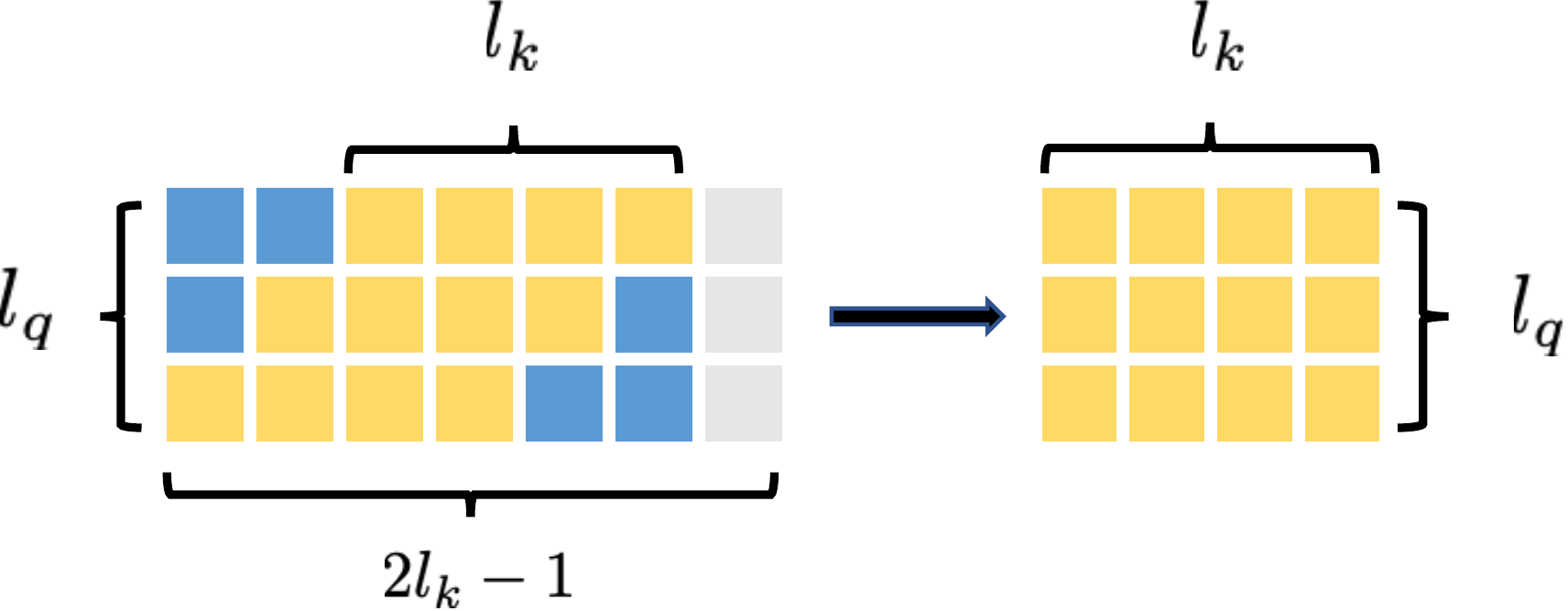}
  \caption{The process of selecting relevant cells from the matrix $\M \in \mathbb{R}^{l_q \times (2l_k-1)}$ (left) and rearranging them into $\M' \in \mathbb{R}^{l_q \times l_k}$ (right). The relevant cells are in yellow, and others are unselected. Note that the position of yellow block in one row of $\M$ is left shifted by 1 cell from the yellow block in the row above.}
  \label{fig:relative_to_absolute}
  \vspace{-4mm}
\end{figure}

Let's denote the length of the query and the extended key as $l_q$ and $l_k$, respectively, where $l_q < l_k$. There are $(2l_k-1)$ possible relative positions from the query to the key ranging in $[-l_k+1,l_k-1]$. Given an embedding matrix $\mathbf{E} \in \mathbb{R}^{d_h \times (2l_k-1)}$, we first compute its dot product with the query $\mathbf{Q}$, resulting in a matrix $\M = \mathbf{Q}^\top \mathbf{E} \in \mathbb{R}^{l_q \times (2l_k-1)}$. Then for the $i$-th row in $\M$, we select $l_k$ consecutive elements corresponding to $l_k$ different relative positions from the $i$-th frame in the query to each frame in the key, and rearrange them into  $\M' \in \mathbb{R}^{l_q \times l_k}$. This process is illustrated in Fig. \ref{fig:relative_to_absolute}. In the $0$-th row, we keep those corresponding to the relative positions in the range $[-l_k+l_q,l_q-1]$; in the $i$-th row, the range is left shifted by 1 from the one in the $(i-1)$-th row; finally in the $(l_q-1)$-th row, the range would be $[-l_k+1,0]$. This process can be conveniently implemented by reusing most of the memory configuration from $\M$ for $\M'$ without copying the underlying storage of $\M$, and then do the following steps: 1) point $\M'$ to the position of the first yellow cell in $\M$; 2) modify the row stride of $\M'$ from $l_k$ to $(l_k-1)$; and 3) modify the number of columns of $\M'$ from $(2l_k-1)$ to $l_k$.
 
\vspace{-5.5mm}
\section{Experiments}
\vspace{-2.5mm}
\subsection{The Dataset}
\vspace{-1.5mm}
We use the Mobvoi (SLR87) dataset\footnote{~\url{https://www.openslr.org/87}} \cite{hou2019region} including two wake words: ``Hi Xiaowen'' and ``Nihao Wenwen''. It contains 144 hrs training data with 43,625 out of 174,592 positive examples, and 74 hrs test data with 21,282 out of 73,459 positive examples. We do not report results on the other datasets mentioned in \cite{wang2020wake}, because both the numbers reported there and in our own experiments are too good (FRR $<0.1$\%) to demonstrate any significant difference.

\vspace{-3mm}
\subsection{Experimental Settings}
\vspace{-1.5mm}
All the experiments in this paper are conducted in \espresso, a PyTorch-based end-to-end ASR toolkit \cite{wang2019espresso}, using \pychain, a fully parallelized PyTorch implementation of LF-MMI \cite{shao2020pychain}.

We follow exactly the same data preparation and preprocessing pipeline as those in \cite{wang2020wake}, including HMM and decoding graph topolopies, feature extraction, negative recording sub-segmentation, and data augmentations. During evaluation, when one of the two wake words is considered, the other one is treated as negative. The operation points are obtained by varying the positive path cost while fixing the negative path cost at 0 in the decoding graph. It it worth mentioning that all the results reported here are from an offline decoding procedure, as currently Kaldi \cite{povey2011kaldi} does not support online decoding with PyTorch-trained neural acoustic models. However, we believe that the offline decoding results would not deviate significantly from the online ones.

The baseline system is a 5-layer dilated 1D convolution network with 48 filters and the kernel size of 5 for each layer, leading to 30 frames for both left and right context (25\% less than that in \cite{wang2020wake}) and only 58k parameters (60\% less than that in \cite{wang2020wake}). For the streaming Transformer models, the first two layers are 1D convolution. They are then followed by 3 self-attention layers with the embedding dimension 32 and the number of heads 4, resulting in 48k parameters without any relative embedding\footnote{~See Table \ref{tab:transformer_cache} for model sizes with different relative embedding settings.}. To make sure that the outputs can ``see'' approximately the same amount of context as those in the baseline, the chunk size is set to 27, so that in the no state-caching setting the right-most frame in a chunk depends on 27 input frames (still smaller than 30) as its right context \footnote{~Our experiments (not shown here) also suggest 27 is the optimal in this setting: a smaller chunk hurts the performance, and a larger one does not have significantly improvement but incurs more latency.}; in the state-caching case, the receptive field covers one more chunk (or 27 more frames) on the left, as it increases linearly when the self-attention layers increases.

All the models are optimized using Adam with an initial learning rate $10^{-3}$ , and then
halved if the validation loss at the end of an epoch does not improve over the previous epoch. The training process stops if the number of epochs exceeds 15, or the learning rate is less than $10^{-5}$. We found that learning rate warm-up is not necessary to train our Transformer-based systems, probably due to the relatively simple supervisions in our task.

\vspace{-2.5mm}
\subsection{Streaming Transformers with State-caching}
\vspace{-1mm}
We first evaluate our streaming Transformer models with state-caching. The results are reported in Table \ref{tab:transformer_cache}, as false rejection rate (FRR) at 0.5 false alarms per hour (FAH). If we only rely on the current chunk and the cached state from the previous chunk but without taking any look-ahead to the future chunk, the detection results (see row 2 in Table \ref{tab:transformer_cache}) are much worse than the baseline. It is actually expected, as the symmetric property of convolution kernels allows the network to take future frames into consideration. This validates that look-head to the future frames is important in the chunk-wise training of Transformers. Then adding absolute positional embedding seems not improve the performance significantly. One possible explanation could be: the goal of the wake word detection is not trying to transcribe the whole recording, but just spot the word of interest, where the absolute encoding of positions do not have too much effective impact. On contrary, when we add relative positional embedding to the key of self-attention layers instead, there is slightly improvement over adding the absolute embedding, which supports our previous hypothesis that the relative embedding makes more sense in such task. When the embedding is also added to the value, FRR reaches 0.7\% and 0.5\% at FAH=0.5 for the two wake words respectively (i.e., 25\% relative improvement over the baseline on average), showing that the embedding is not only useful when calculating the attention weights, but also beneficial when encoding the positions into the layer's hidden values.
\begin{table}
  \caption{Results of streaming Transformers with state-caching.}
  \vspace{-4mm}
  \begin{center}
    \begin{adjustbox}{max width=\linewidth}
    \begin{tabular}{ l c c c}
    \toprule
      & \multirow{2}{*}{\#Params} & \multicolumn{2}{c}{FRR(\%) at FAH=0.5} \\
    \cmidrule(lr){3-4}
     & & Hi Xiaowen & Nihao Wenwen\\
       1D Conv. (baseline)\tablefootnote{~We do not compare with other systems, because to our best knowledge this baseline system is the state-of-the-art reported on the same dataset at the time of submission.}  & 58k & 0.8 & 0.8 \\
       Transformer (w/o look-ahead) & 48k &  3.5 &  4.7 \\
       \quad+look-ahead to next chunk & 48k & 1.3 &  1.2 \\
       \quad\quad+abs. emb. & 48k & 1.2 &  1.2 \\
       \quad\quad+rel. emb. to key & 52k &  1.0 &  1.1 \\
       \quad\quad\quad+rel. emb. to value & 57k &  \textbf{0.7} &  \textbf{0.5} \\
    \bottomrule
    \end{tabular}
    \end{adjustbox}
  \end{center}
  \label{tab:transformer_cache}
  \vspace{-8mm}
\end{table}

\vspace{-2.5mm}
\subsection{Streaming Transformers without State-caching}
\vspace{-1mm}
Next we would like to explore whether having gradient to been back-propagated into the history state would help train a better model. As we mentioned in Sec. \ref{sec:look-ahead-and-caching}, this can be done by concatenating the current chunk with the previous chunk of input, instead of caching the internal state of the previous chunk. Table \ref{tab:transformer_nocache} shows several results. By looking at Table \ref{tab:transformer_nocache} itself, we observe a similar trend as that in the state-caching model from the previous section: relative positional embedding is advantageous over the absolute sinusoidal positional embedding, and adding the embedding to both key and value is again the best. Furthermore, by comparing the rows in Table \ref{tab:transformer_nocache} with their corresponding entries in Table \ref{tab:transformer_cache}, we observe that, except the case in the last row, regardless of the choice of positional embedding and how it is applied, the models without state-caching outperform models with state-caching. It indicates the benefit of updating the model parameters with more gradient information back-propagated from the current chunk into the previous chunk. However in the case where relative positional embedding is also added to the value, the gap seems diminished, suggesting that by utilizing the positional embedding in a better way, there is no need to recompute the part of the cached state in order to reach the best performance.

\begin{table}
  \caption{Results of streaming Transformers without state-caching.}
  \vspace{-4mm}
  \begin{center}
    \begin{adjustbox}{max width=\linewidth}
    \begin{tabular}{ l c c c}
    \toprule
      & \multirow{2}{*}{\#Params} & \multicolumn{2}{c}{FRR(\%) at FAH=0.5} \\
    \cmidrule(lr){3-4}
     & & Hi Xiaowen & Nihao Wenwen \\
       1D Conv. (baseline)  & 58k & 0.8 & 0.8 \\
       Transformer (w/ look-ahead) & 48k & 1.0 &  1.1 \\
       \quad+abs. emb. & 48k &  0.8 &  0.8 \\
       \quad+rel. emb. to key & 52k &  0.6 &  0.7 \\
       \quad\quad+rel. emb. to value & 57k &  \textbf{0.6} &  \textbf{0.6} \\
    \bottomrule
    \end{tabular}
    \end{adjustbox}
  \end{center}
  \label{tab:transformer_nocache}
  \vspace{-8mm}
\end{table}

\begin{figure}[ht]
  \centering
  \includegraphics[width=0.37\textwidth]{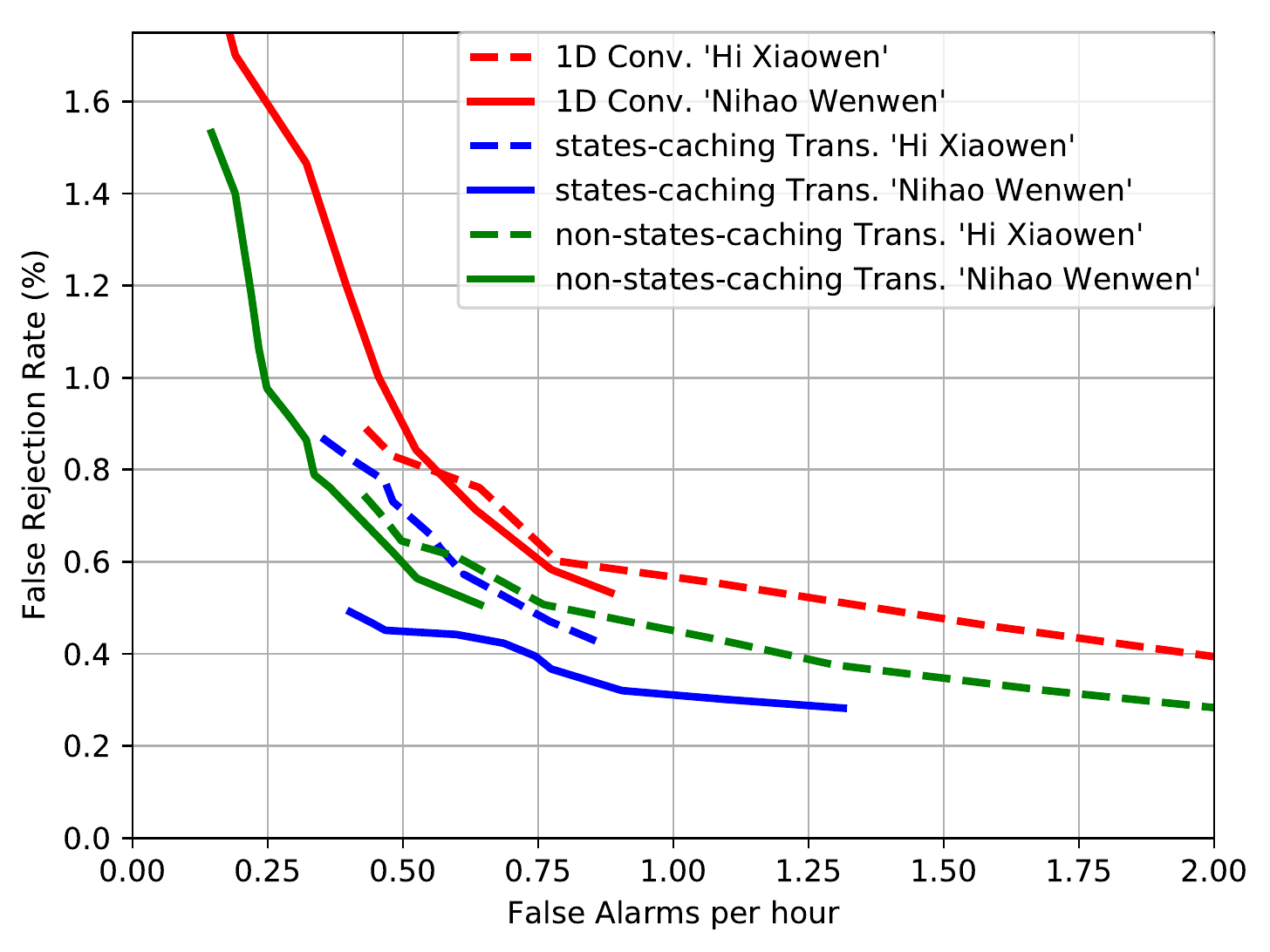}
  \caption{DET curves for the baseline 1D convolution network and our two proposed streaming Transformers.}
  \label{fig:det}
\end{figure}

We provide DET curves of the baseline convolution network and the two proposed streaming Transformers in Fig. \ref{fig:det}, for a more comprehensive demonstration of their performance difference.

\vspace{-2mm}

\vspace{-6.5mm}
\subsection{Streaming Transformers with Same-layer Dependency}
\vspace{-1mm}
We now explore the architectural variant introduced in Sec. \ref{sec:same-layer}. Note that if the relative positional embedding is added to the value $\mathbf{V}_{c-1}^l$ as shown in Eq. (\ref{eq:compute_val_pos_embed}), $\H_{c-1}^l$ will no longer be in the same semantic space as $\mathbf{V}_c^l$. So it is problematic to concatenate $\H_{c-1}^l$ and $\mathbf{V}_c^l$ together in Eq. (\ref{eq:state_extended_samelayer}). A similar issue arises if the parameter $\W_K$ and $\W_V$ from the same layer are not tied because $\H_{c-1}^l$ is going to be concatenated to both $\mathbf{K}_c^l$ and $\mathbf{V}_c^l$. Our solution is to only add the positional embedding to $\mathbf{K}_c^l$, and also tie $\mathbf{K}_c^l$ and $\mathbf{V}_c^l$ together. However, it only achieves FRR=1.3\% at FAH=0.5. When absolute embedding is used, FRR=1.1\% at the same FAH. This contradicts the observations in \cite{lu2020exploring,wu2020streaming} where same-layer dependency was found to be more advantageous for ASR and it was attributed to the fact that the receptive field is maximized at every layer\footnote{~They did not mention the type of positional embedding being used.}. A better way of incorporating relative positional information for this case is our future work.

\vspace{-3mm}
\section{Conclusions}
\vspace{-2mm}
We propose using streaming Transformers for wake word detection with the latest alignment-free LF-MMI system. We explore how look-ahead of the future chunk, and different gradient stopping, layer dependency, and positional embedding strategies could affect the system performance. Along the way we also propose a series of simple tensor operations to efficiently compute the self-attention in the streaming setting when relative positional embedding is involved. Experiments on Mobvoi (SLR87) show the advantage of the proposed streaming Transformers over the 1D convolution baseline.

\bibliographystyle{IEEEbib}
\fontsize{8.8}{10.3}\selectfont
\bibliography{refs}

\end{document}